\documentclass{article} 

\usepackage{amsmath,amsfonts,bm}

\def\eqref#1{equation~\ref{#1}}

\def\1{\bm{1}}

\DeclareMathAlphabet{\mathsfit}{\encodingdefault}{\sfdefault}{m}{sl}
\SetMathAlphabet{\mathsfit}{bold}{\encodingdefault}{\sfdefault}{bx}{n}

\usepackage{hyperref}
\usepackage{url}

\usepackage{hyperref}       
\usepackage{url}            
\usepackage{booktabs}       
\usepackage{amsfonts}       
\usepackage{nicefrac}       
\usepackage{microtype}      
\usepackage{xcolor}         
\usepackage{graphicx}
\usepackage{cleveref}
\usepackage{multirow} 
\usepackage{tabularx}
\usepackage{enumitem}
\usepackage{float}
\usepackage{subcaption}
\usepackage{natbib}
\usepackage{times}

\title{Llamba: Scaling Distilled Recurrent Models for Efficient Language Processing}

\usepackage{authblk}

\author[$^{1}$]{Aviv Bick\thanks{Work was done while at Cartesia AI.}}
\author[$^2$]{Tobias Katsch}
\author[$^2$]{Nimit Sohoni}
\author[$^2$]{Arjun Desai}
\author[$^{12}$]{Albert Gu}

\affil[$^1$]{Carnegie Mellon University}
\affil[$^2$]{Cartesia.ai}

\affil[ ]{{\texttt{abick@cs.cmu.edu}}}

\date{}

\begin{document}

\maketitle

\begin{abstract}
We introduce Llamba, a family of efficient recurrent language models distilled from Llama-3.x into the Mamba architecture.
The series includes Llamba-1B, Llamba-3B, and Llamba-8B, which achieve higher inference throughput and handle significantly larger batch sizes than Transformer-based models, while maintaining comparable benchmark performance.
Furthermore, Llamba demonstrates the effectiveness of cross-architecture distillation using MOHAWK \citep{mohawk}, achieving these results with less than 0.1\% of the training data typically used for models of similar size.
To take full advantage of their efficiency, we provide an optimized implementation of Llamba for resource-constrained devices such as smartphones and edge platforms, offering a practical and memory-efficient alternative to Transformers.
Overall, Llamba improves the tradeoff between speed, memory efficiency, and performance, making high-quality language models more accessible.
\end{abstract}

\section{Introduction}

Transformer-based LLMs dominate language modeling, but their quadratic attention mechanism makes them computationally expensive and difficult to scale efficiently.
This technical paper introduces the \textbf{Llamba model family}, a suite of SSM-based language models—including Llamba-1B, Llamba-3B, and Llamba-8B—that address these efficiency challenges.
Retaining the overall structure of Llama models, Llamba models are distilled from Llama-3, replacing self-attention with Mamba-2 layers to achieve high inference throughput while maintaining strong performance across benchmarks.

Llamba achieves its performance with drastically reduced training data requirements through \emph{architecture distillation}.
Unlike traditional large language models (LLMs) that rely on massive datasets spanning trillions of tokens, Llamba achieves comparable results with significantly fewer resources by using MOHAWK \citep{mohawk} to transfer the knowledge from strong pretrained Transformer-based LLMs 
to a Mamba-based architecture.
For example, \textit{Llamba-3.1-8B was distilled using just 12 billion tokens—less than 0.1\% of the training data required for Llama-3.1-8B}.
This remarkable data efficiency highlights the effectiveness of architecture distillation methods
in transferring knowledge from pretrained models while significantly reducing both data and computational demands. As a result, Llamba presents a scalable and cost-effective solution for high-performance language modeling.

Extending the benefits of their efficient design, \textbf{we provide on-device implementations of the Llamba models}
\footnote{https://github.com/cartesia-ai/edge}, optimized for deployment on private devices such as smartphones and edge computing platforms with limited memory and computational resources. 
These implementations highlight the advantages of linear models, such as the Llamba family, by delivering high-quality language modeling on devices where traditional Transformer architectures are often impractical due to their heavy memory and compute demands.

Overall, the Llamba family introduces several key contributions:
\begin{itemize}[leftmargin=*]
    \item \textbf{Distillation efficiency:} Using the MOHAWK framework, Llamba achieves state-of-the-art performance with less than 0.1\% of the training data required by comparable models. This represents a significant advancement in data and compute efficiency for LLMs.
    \item \textbf{On-device deployment:} We provide quantized Llamba models, along with an MLX implementation for edge devices like iPhones and MacBooks, which demonstrate near-constant memory usage, making them ideal for resource-constrained environments.
    \item \textbf{Benchmark performance:} Llamba-1B, Llamba-3B, and Llamba-8B perform on par with traditional models across a wide range of benchmarks, setting a new standard for efficiency and performance in recurrent architectures.
\end{itemize}

These advancements position Llamba as a versatile and scalable solution for efficient language modeling, bridging the gap between performance, resource efficiency, and accessibility.

\begin{figure}[!t]
    \centering
    \includegraphics[width=0.9\linewidth]
    {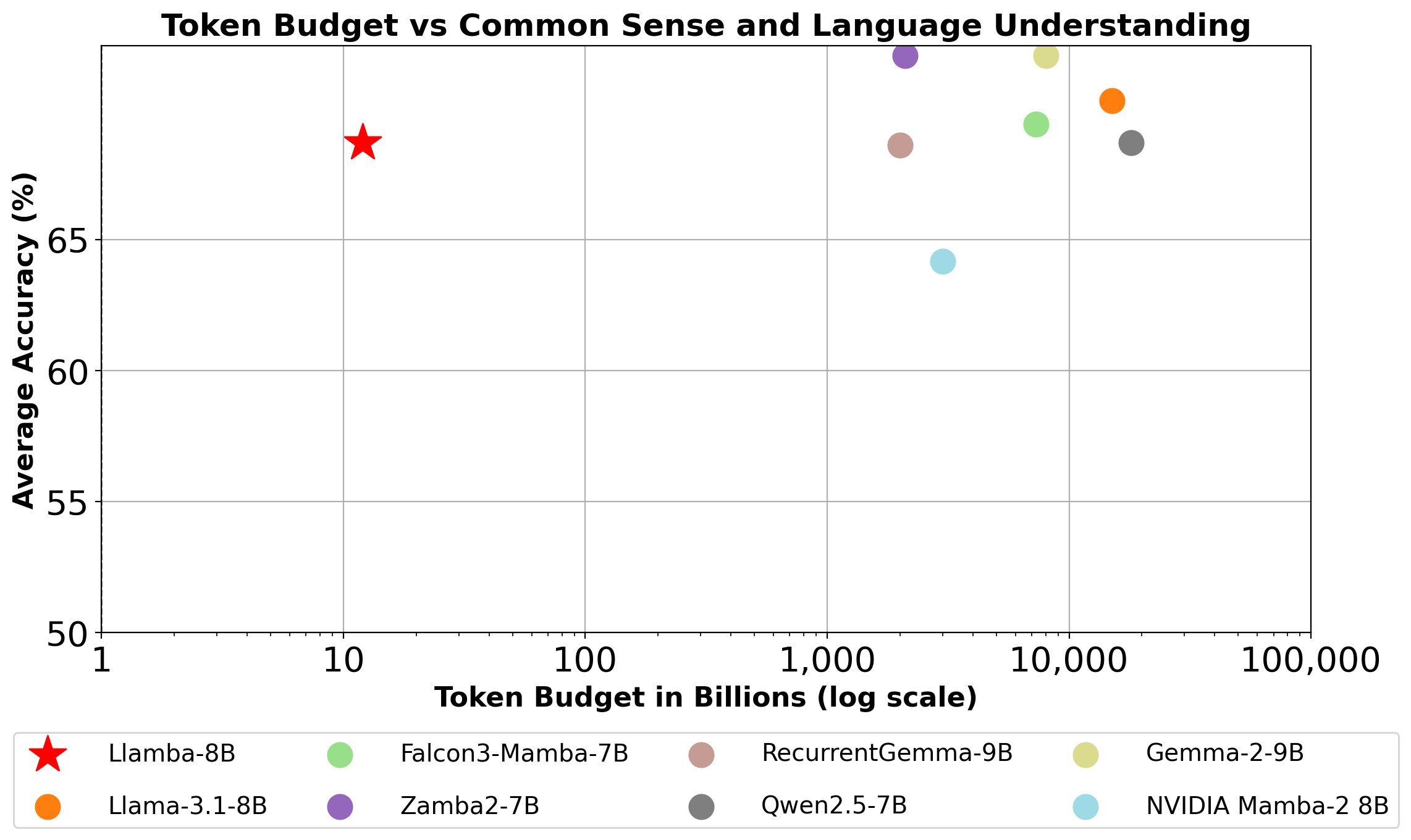}
    \caption{
    Average accuracy is measured over multiple benchmarks, including ARC Challenge, ARC Easy, PIQA, Winogrande, HELLASWAG, OpenBookQA, and MMLU, providing a comprehensive assessment of a model's Common Sense and Language Understanding.
    }
    \label{fig:tokens_budget}
\end{figure}

\begin{table}[H]
\centering
\small
\setlength{\tabcolsep}{3.1pt}
\caption{
Comparison of downstream performance (accuracy \%) in zero-shot settings across various models. 
We use the following abbreviations: 
ARC Challenge (ARC-C), 
ARC Easy (ARC-E), 
Physical Interaction QA (PIQA), 
Winogrande (WG), 
HellaSwag (HS), 
Lambada (LMB), 
Massive Multitask Language Understanding (MMLU), 
and OpenBookQA (OBQA). 
For ARC-C, ARC-E, PIQA, HellaSwag, and OBQA, we use normalized logits' results. 
Averages are computed over all 8 tasks. 
Along with accuracy, each model is annotated with the number of training or distillation tokens (in trillions) and its architecture—Recurrent (R), Transformer (T), or Hybrid (H). For models with a sliding window, the window size is also specified.
This table shows that all Llamba models achieve competitive performance against both Transformer-based and SSM-based models, despite their recurrent layers being trained on significantly fewer tokens.
We use an instruct-tuned version whenever one is available; however, we exclude this label for brevity.
See \Cref{tab:performance_comparison_1,tab:performance_comparison_2} for the full few-shot comparisons.
}
\label{tab:performance_comparison}
\begin{tabular}{l c c c c c c c c c c c}
\toprule
\textsc{Model} & 
\textsc{Arch.} & 
\textsc{Tokens (T)} & \textsc{ARC-C} & 
\textsc{ARC-E} & \textsc{PIQA}  & 
\textsc{WG} & \textsc{HS} & 
\textsc{LMB} & \textsc{MMLU} & 
\textsc{OBQA} & \textbf{AVG} \\

\midrule

\textsc{Llama-3.2-1B (Teacher)} & T
& $\le$ 9   & 38.1 & 68.5 & 74.4 & 59.7 & 60.8 & 60.1 & 46.0 & 34.6 & 55.3 \\

\textbf{\textsc{Llamba-1B}} & R
& 0.008     & 37.2 & 69.5 & 74.0 & 60.6 & 61.2 & 48.4 & 38.0 & 37.0 & 53.2 \\

\textsc{RecurrentGemma-2B} & H ($w=2048$)
& $\le$ 2   & 35.6 & 51.2 & 67.2 & 55.7 & 60.3 & 52.5 & 40.2 & 30.4 & 49.1 \\

\midrule

\textsc{Qwen2.5-3B} & T
& $\le$ 18  & 48.1 & 72.9 & 78.3 & 69.8 & 74.9 & 65.8 & 65.5 & 41.8 & 64.6 \\

\textsc{Llama-3.2-3B (Teacher)} & T
& 9         & 45.6 & 74.3 & 75.8 & 67.6 & 70.4 & 65.9 & 60.4 & 35.8 & 61.9 \\

\textbf{\textsc{Llamba-3B}} & R
& 0.01      & 48.5 & 79.0 & 78.6 & 70.4 & 73.8 & 65.8 & 52.7 & 42.8 & 63.9 \\

\textsc{Mamba2-2.8B} & R
& 0.3       & 35.9 & 64.3 & 75.6 & 63.4 & 66.2 & 68.1 & 25.7 & 40.4 & 54.9 \\

\midrule

\textsc{Qwen2.5-7B} & T
& 18        & 55.1 & 81.3 & 80.3 & 71.1 & 80.5 & 69.5 & 71.7 & 48.6 & 69.8 \\

\textsc{Llama-3.1-8B (Teacher)} & T
& 15        & 55.1 & 81.7 & 81.1 & 73.9 & 79.3 & 73.0 & 68.0 & 43.0 & 69.4 \\

\textbf{\textsc{Llamba-8B}} & R
& 0.012     & 54.6 & 82.5 & 80.9 & 73.3 & 77.6 & 69.4 & 61.0 & 43.4 & 68.8 \\

\textsc{Falcon3-Mamba-7B} & R
& 7.3       & 53.2 & 72.5 & 79.7 & 69.1 & 79.8 & 67.5 & 65.0 & 48.0 & 67.0 \\

\textsc{Zamba2-7B} & H
& 2.1       & 56.1 & 80.6 & 81.1 & 76.9 & 81.5 & 74.6 & 64.7 & 45.2 & 70.1 \\

\textsc{RecurrentGemma-9B} & H ($w=2048$)
& 2         & 57.1 & 78.9 & 80.6 & 73.7 & 80.1 & 54.1 & 55.1 & 46.0 & 65.7 \\

\bottomrule
\end{tabular}
\end{table}
\section{Related Work}

\paragraph{Language Models.}
Transformer-based models, such as those in the Llama \citep{llama}, and Qwen \citep{qwen2} series, have shown strong performance across various language modeling tasks. These models underwent extensive pretraining on large-scale corpora and incorporate techniques like instruction tuning and curated datasets to enhance generalization in few-shot and zero-shot settings on various tasks.

While Transformers remain dominant, recent work has explored alternatives to their purely quadratic attention mechanisms to improve efficiency while maintaining strong performance. Structured state space models (SSMs) \citep{mamba1, mamba2} have emerged as a promising direction, offering a distinct approach to sequence modeling.
At large scales, Falcon-Mamba \citep{falcon}, a fully SSM-based model stacking Mamba-1 layers, has matched and even outperformed Transformers on key tasks. Falcon3-Mamba extends this by pretraining for an additional 1.5 trillion tokens, incorporating high-quality post-training data, and expanding the context length from 8K to 32K tokens, further enhancing its capabilities.
However, despite these advances, SSM-based models still underperform Transformers on algorithmic tasks \citep{repeat_after_me,wen2024rnnstransformersyetkey,waleffe2024}.

To balance these trade-offs, hybrid models combining attention and SSMs have gained interest for leveraging the strengths of both architectures.
Examples include RecurrentGemma \citep{recurrentgemma}, which integrates gated linear recurrences with local attention, and Zyphra’s Zamba \citep{zamba}, which combines Mamba-1 layers with a shared attention mechanism spanning the network.
Zamba-2 \citep{zamba2} builds on this by replacing Mamba-1 blocks with Mamba-2 for improved efficiency, increasing shared attention layers from one to two for enhanced global context modeling, and applying Low-Rank Adaptation (LoRA) \citep{lora} to shared MLP blocks for parameter-efficient depth adjustments.
Other hybrid architectures \citep{jamba2024, waleffe2024} further underscore the interest in models that balance expressiveness and efficiency.

\paragraph{Distillation.}
Various methods have been proposed to distill large Transformer-based models into more efficient architectures while maintaining performance.
SUPRA \citep{mercat2024linearizing} propose a procedure to linearize softmax attention into a form of linear attention by copying the weight matrices and fine-tuning.
LoLCATs \citep{lolcats} introduces a linearization approach that replaces softmax attention with linear attention through attention transfer, followed by low-rank adaptation, enabling the creation of large-scale linearized models with improved efficiency. 
\citep{wang2025} leverages the State-Space Duality (SSD) in \citet{mamba2} to transfer the linear projection weights from the attention layers into Mamba-based models. Their experiments include hybrid models with an increasing proportion of interleaved attention layers, demonstrating that a balanced combination of state-space models (SSMs) and attention preserves performance while improving efficiency.
MOHAWK \citep{mohawk} distills Transformers into SSMs through progressive alignment, allowing subquadratic models to leverage Transformer training resources effectively. These approaches demonstrate the viability of distilling computationally expensive Transformers into efficient models while retaining competitive performance.
\section{Model Architecture}
\label{model_architecture}

\begin{figure}[t!]
    \centering
    \begin{subfigure}[b]{0.4\linewidth}
        \centering
        \includegraphics[width=0.7\linewidth]{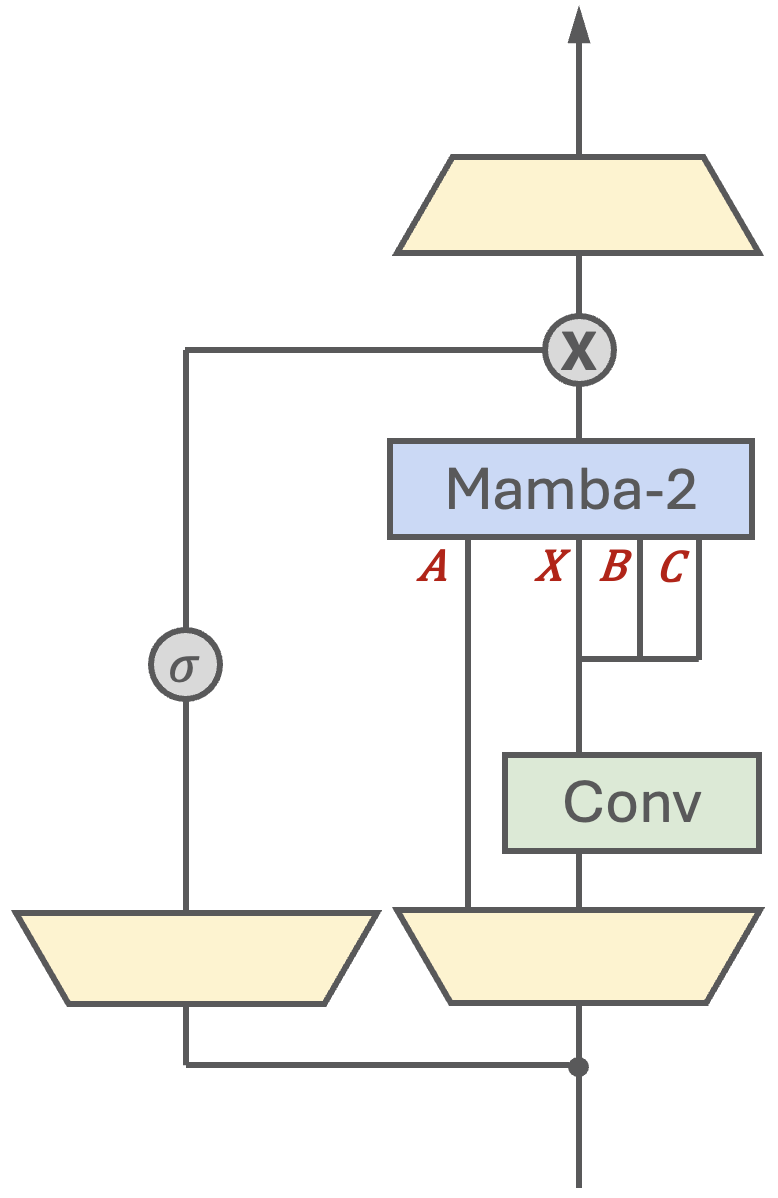}
        \caption{
        The Discrete Mamba-2 block \cite{mohawk} modifies the original Mamba-2 architecture by removing both post-convolution activation and pre-output projection normalization. Additionally, the Discrete Mamba-2 sequence mixer eliminates the $\Delta$ discretization parameter and directly projects the $\mathbf{A}$ matrix from the input.
        }
        \label{fig:discrete_mamba}
    \end{subfigure}
    \hspace{0.1\linewidth} 
    \begin{subfigure}[b]{0.4\linewidth}
        \centering
        \includegraphics[width=0.62\linewidth]{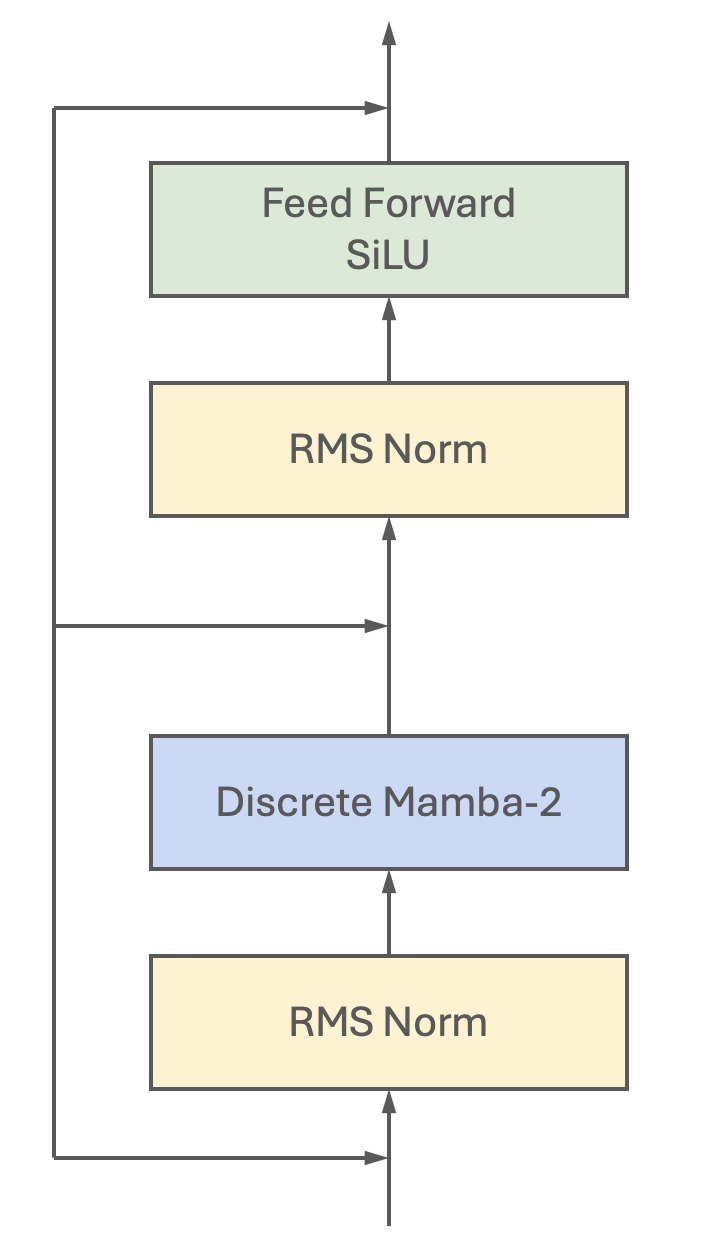}
        \caption{Llamba models—Llamba-1B, Llamba-3B, and Llamba-8B—are based on the architecture of their Llama teacher models. Each block comprises two sub-blocks with residual connections:
        (1) RMS Normalization followed by a Discrete Mamba-2 layer.
        (2) RMS Normalization followed by a feed-forward layer.
        }
        \label{fig:llamba_architecture}
    \end{subfigure}
    \caption{Comparison of the Discrete Mamba-2 block and the Llamba architecture.}
    \label{fig:comparison}
\end{figure}

Unlike the Mamba and Mamba-2 architectures, which were designed for training from scratch, \textit{Llamba is directly motivated by architectural distillation}.
In particular, the Mohawk distillation framework involves aligning sub-networks of the model at various levels of granularity (\Cref{sec:distillation}). 
This constraints Llamba to retain the overall architecture of the teacher model, ideally modifying only the attention matrix mixer by replacing it with a subquadratic alternative.

The Llamba models—Llamba-1B, Llamba-3B, and Llamba-8B—comprise 16, 28, and 32 residual Mamba-2 blocks, respectively, followed by feed-forward layers. These models share the tokenizer and vocabulary of Llama-3.1, with hidden dimensions of 2048 for Llamba-1B, 3072 for Llamba-3B, and 4096 for Llamba-8B. 
In addition, Llamba differs from the original Mamba-2 architecture \citep{mamba2} in the following ways (see \Cref{fig:llamba_architecture}):
\begin{itemize}[leftmargin=*]

\item \textbf{Alternating MLP blocks}:
Llamba interleaves Llama’s Gated MLP components between each Mamba-2 mixing layer, unlike Mamba-1 and Mamba-2, which consist solely of SSM blocks.

\item \textbf{Multi-head structure}: Llama-3.x models use grouped-query attention (GQA) \citep{gqa, mqa}, which employs 32 query heads and 8 key-value heads to boost inference speed and reduce the size of the decoding cache. However, Mamba’s recurrent layers don’t rely on a cache, so these optimizations aren’t needed. Instead, \textit{Llamba blocks feature a Multi-Head variant} of Mamba-2 with 32 heads and dimensions of 64, 96, or 128, along with a state size of 64. While this design differs from Mamba-2’s ``multi-value attention'' (MVA) architecture, it still keeps inference costs low.

\item \textbf{Non-linearities}: We remove the normalization before the output projection and the activation after convolution, as these are non-linear operations that do not exist in the attention block and hurts alignment (See \Cref{subsec:mohawk}). 

\item \textbf{Discretization}: Llamba uses \textit{Discrete-Mamba-2}, a variant that projects the matrix $\mathbf{A}$ directly from the input, eliminating the discretization parameters $\Delta$ to better match the inherently discrete attention mechanisms. 
\end{itemize}

Notably, these changes not only facilitate the distillation process but also improve training efficiency. Alternating with MLPs \textbf{reduces the number of temporal mixing layers}, enabling Llamba to achieve faster computation than other models of comparable size (see \Cref{subsec:throughput}). Furthermore, training becomes simpler and more efficient by eliminating normalization-related all-reduce operations.

\section{Distillation}
\label{sec:distillation}

The Llamba models were distilled using MOHAWK \citep{mohawk} from Meta’s Llama-3.x family \citep{llama}. Specifically, Llamba-3.1-1B was distilled from Llama-3.2-1B-Instruct, Llamba-3B from Llama-3.2-3B-Instruct, and Llamba-8B from Llama-3.1-8B-Instruct.

\subsection{MOHAWK}
\label{subsec:mohawk}

Following the MOHAWK framework \citep{mohawk}, Llamba models were initialized by setting the convolution layer of the Mamba block to an identity kernel (nullifying its effect) and configuring the multiplicative skip connection to directly pass the input unchanged, effectively initializing the block as an identity function.
Subsequently, three key steps were implemented: \textit{Matrix Orientation}, \textit{Hidden-State Alignment}, \textit{Weight Transfer with Knowledge Distillation}.

\paragraph{Matrix Orientation.}
This phase is used to align the student and teacher matrix mixers. Specifically, we minimize the distance between the materialized Llamba matrix mixer and Llama’s self-attention matrix. Notably, Llama uses an MQA architecture, which results in 32 attention heads with shared weights. Since Llamba’s 32 heads are not tied (it uses a Multi-Head architecture), it learns independent weights, unlike the dependent matrices of its teacher.

\paragraph{Hidden-State Alignment.}
For Hidden-State Alignment, each Mamba-2 block of the Llamba model was aligned independently using the L2 distance, guided by the output of the preceding layer.

\paragraph{Weight Transfer \& Knowledge Distillation.}
We begin this stage by transferring the MLP weights, normalization layers, input embedding, and output head from the Llama-3.x models to each Llamba model.
Unlike previous works \citep{wang2024mamballamadistillingaccelerating,mohawk}, we did not freeze the MLP components and optimized them using the same learning rate of the mixing components.
During Knowledge Distillation, each Llamba model was aligned with the respective teacher model using the cross-entropy loss of their logits. After this phase's loss saturation, all models were further distilled from Llama-3.1-70B-Instruct for their remaining tokens.

\subsection{Training Details}
\label{subsec:training_details}

\begin{table}[!t]
\centering
\begin{tabular}{@{}lllll@{}}
\toprule
  & \textsc{Stage 1} & \textsc{Stage 2} & \textsc{Stage 3} & \textsc{Overall Tokens} \\
  \midrule
\textsc{Llamba-1B} & 300M & 2.7B & 5B & 8B \\
\textsc{Llamba-3B} & 500M & 4B & 5.5B & 10B \\
\textsc{Llamba-8B} & 500M & 5B & 6.5B & 12B \\
\bottomrule
\end{tabular}
\caption{
Token allocations during the distillation process for different Llamba models and MOHAWK stages (Matrix Orientation, Hidden-State Alignment, and Knowledge distillation).
}
\label{tab:stages_tokens_budget}
\end{table}

The Llamba models were trained using mixed precision and Fully Sharded Data Parallel (FSDP) on a single node with 8 H100 GPUs, with activation checkpointing enabled.
Training used the AdamW optimizer with $\beta_1 = 0.9$, $\beta_2 = 0.95$, and a weight decay of 0.1. The maximum learning rates were $1 \times 10^{-4}$ for the first two MOHAWK stages across all models, $5 \times 10^{-5}$ for the third stage of Llamba-1B and Llamba-3B, and $1 \times 10^{-5}$ for the third stage of Llamba-8B. Batch sizes were set to $64$ in the first MOHAWK stage and $128$ in the second and third stages.
We used the Warm-Stable-Decay (WSD) scheduler \citep{minicpm}, with a minimum learning rate of $1 \times 10^{-8}$ and warm-up and decay phases each spanning 10\% of total training steps.

During the distillation process \Cref{tab:stages_tokens_budget}, a total of $12$ billion tokens were processed for Llamba-8B, $10$ billion tokens for Llamba-3B and Llamba-1B used only $8$ billion tokens, highlighting its significantly smaller allocation of training data used for distillation compared to training without any teacher supervision (see \Cref{fig:tokens_budget}).

\subsection{Data}
\label{mohawk:data}

Data quality is critical for accurately modeling temporal interactions in the MOHAWK distillation setting. MOHAWK transfers only the MLP weights that affect the hidden dimensions, excluding the sequence mixer weights related to the time dimension. This omission limits the ability to capture time-dependent information directly.
For the distillation process, two datasets were used. The first, fineweb-edu-4.0, is derived from fineweb-edu \citep{fineweb}, itself a subset of the broader fineweb dataset. This refined subset includes only highly educational web pages, filtered using a 4.0 classifier score threshold - higher than the 3.0 threshold used for fineweb-edu. Since distillation requires relatively few tokens, this focused approach was practical and effective.

The \textit{Matrix Orientation} and \textit{Hidden-State Alignment} processes were conducted using the fineweb-edu-4.0 dataset with packed sequences of length 2048 (see \Cref{tab:stages_tokens_budget} for more details). In contrast, \textit{Knowledge Distillation} was initially performed using fineweb-edu-4.0, and subsequently, the Open-Hermes-2.5 dataset was employed for an additional 4 epochs, processing 200 million tokens per epoch with sequences of length 4096. The combination of these datasets played a pivotal role in enhancing the MMLU score.

\begin{figure}[!t]
    \centering
    \includegraphics[width=0.49\linewidth]
    {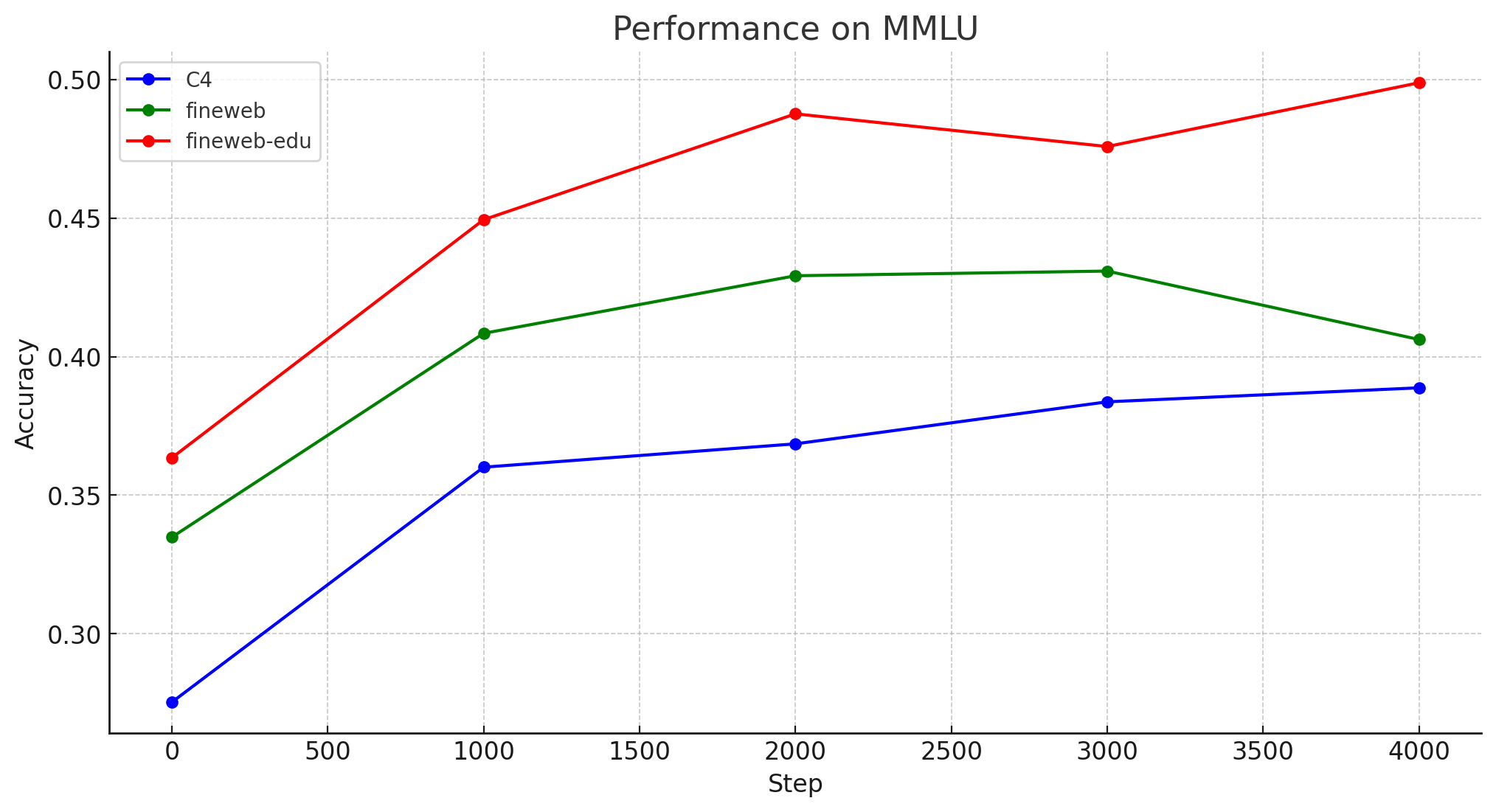}
    \includegraphics[width=0.49\linewidth]
    {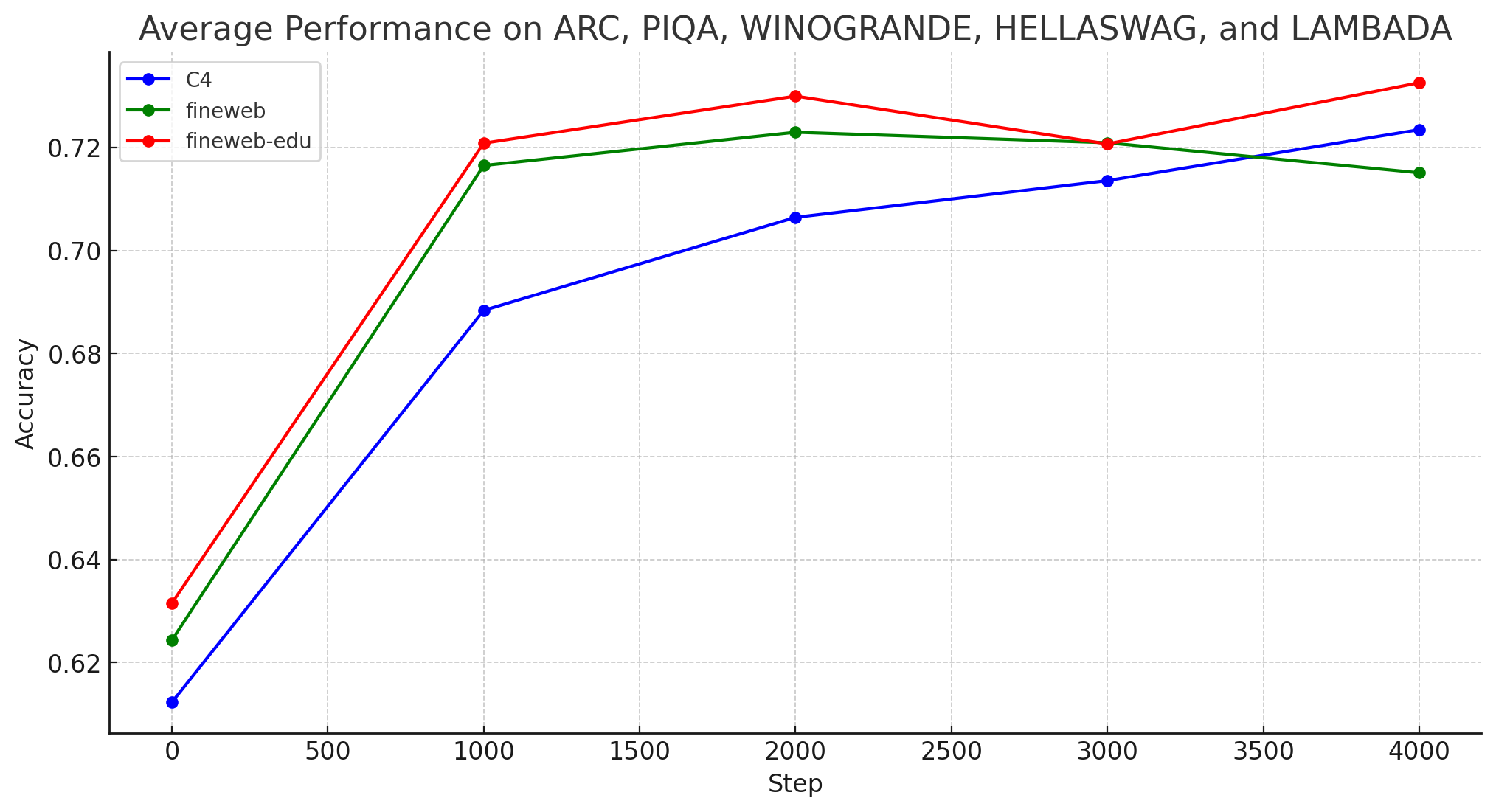}
    \caption{
    An evaluation of Llamba-8B's knowledge distillation step (MOHAWK's stage 3) across three datasets: C4, fineweb, and fineweb-edu. Each model underwent hidden-state alignment (MOHAWK's stage 2) on its respective dataset using 4 billion tokens and subsequently underwent testing with knowledge distillation on 1 billion tokens. It is observed that although all datasets yield similar outcomes across most benchmarks, MMLU shows notable improvement when utilizing fineweb-edu, unlike with fineweb and C4.
    }
    \label{fig:data_comparison}
\end{figure}

Our results demonstrate that this dataset selection significantly improves the performance of MMLU \citep{mmlu}. As shown in \Cref{fig:data_comparison}, while the C4 \citep{C4} and fineweb datasets achieve similar scores across most benchmarks, fineweb-edu drives a marked improvement in MMLU. Beyond this, our approach highlights an important takeaway: \textit{we achieve strong results using only established open-source datasets, in contrast to many alternative models that rely on highly curated proprietary datasets}. This demonstrates the feasibility of leveraging openly available resources for high-quality model performance.

Furthermore, we emphasize that architecture distillation (e.g. the MOHAWK framework) and data curation are orthogonal and synergistic,
and we hypothesize that our distillation results could be improved further by incorporating even higher-quality datasets.

\section{On-Device Implementation}

The advantages of sub-quadratic language models are particularly impactful in compute- and memory-constrained environments, making them ideal for on-device applications. To support efficient inference, we implemented optimized Mamba-2 kernels, including state-space model and Conv1D layers, using Apple’s Metal framework. These kernels are specifically designed for Apple Silicon, leveraging its GPU parallelism and unified memory architecture for efficient execution.

Our implementation integrates seamlessly with MLX \citep{mlx2023}, a machine learning framework optimized for Apple Silicon. MLX enables dynamic graph construction and efficient tensor operations while utilizing unified memory to minimize data transfer overhead. Additionally, we support 4-bit quantization to further reduce memory usage, enabling models to run effectively on devices with limited resources.

These optimizations allow our models to maintain high throughput and low memory consumption, even in long-context scenarios, making them highly suitable for real-time, on-device applications.
The implementation is available in the released repository \url{https://github.com/cartesia-ai/edge}.

\section{Evaluations}

\subsection{Performance}

\paragraph{Comparison Against Pretrained Models.}
\label{subsec:performance}
\Cref{tab:performance_comparison_1} presents a comparative analysis of downstream evaluation results across different models and tasks. The evaluation includes recent advanced models such as hybrids of subquadratic and attention layers (e.g., Zamba2-7B \citep{zamba2}) and purely subquadratic models (e.g., RecurrentGemma-9B \citep{recurrentgemma}, Falcon-Mamba-7B \citep{falcon}). Additionally, we include Llama-3.2-1B, Llama-3.2-3B, and Llama-3.1-8B to benchmark performance against state-of-the-art non-hybrid Transformer models.

We evaluate the models’ performance in both zero-shot and few-shot settings across a range of standard datasets: ARC \citep{arc}, PIQA \citep{piqa}, Winogrande (WG) \citep{winogrande}, HellaSwag (HS) \citep{hellaswag}, Lambada OpenAI (LO) \citep{lambada}, MMLU \citep{mmlu}, and OpenBookQA (OBQA) \citep{OpenBookQA}. 
All evaluations were conducted using \texttt{bfloat16} precision and the \textit{lm-eval-harness v0.4.4} Python library \citep{eval-harness}.

\begin{figure}[t!]
    \centering
    \includegraphics[width=1\linewidth]{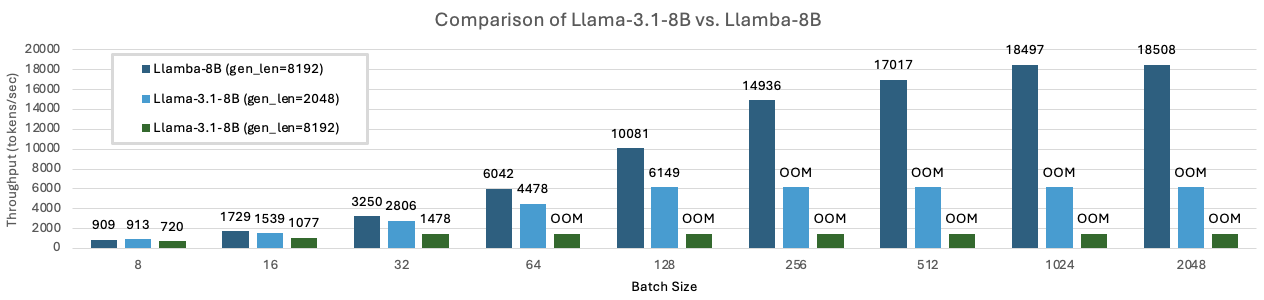}
    \caption{
    Tokens processed at different batch sizes across various models. All models were compiled using \texttt{torch.compile(model, fullgraph=True)} with CUDA graph compilation. We evaluated three settings:
    (1) Llamba-8B with \texttt{gen\_len=8192},
    (2) Llama-3.1-8B with \texttt{gen\_len=2048}, and
    (3) Llama-3.1-8B with \texttt{gen\_len=8192}.
    Each was tested with \texttt{prompt\_len=1} and batch sizes ranging from 8 to 2048. The results show that Llamba-8B achieves the highest throughput, particularly at larger batch sizes, where Transformers either slow down or run out of memory (OOM).
    }
    \label{fig:throughput_over_batch_size}
\end{figure}

\paragraph{Comparison Against Distilled Models.}
\Cref{tab:comparison_against_distillation_methods} compares Llamba-8B with other distilled models of similar size.
We specifically focus on MMLU, which is known to be difficult for recurrent models \citep{waleffe2024}, and has the biggest gap for distilled models from prior work. Llamba significantly improves MMLU relative to the teacher model.

We found that MMLU performance takes much longer to improve compared to other benchmarks in our end-to-end distillation. Llamba reached teacher performance on other tasks in a very small number of tokens, while MMLU took longer to improve. 

We also note that some of the baselines are actually hybrid models, which have a 1:1 ratio of attention to recurrent layers. We note that even sliding window attention has a strong effect because the MMLU context size is very small. 
Although Llamba still has a gap to the teacher model, we consider this result a large step forward for the performance of distilled recurrent models.


\begin{table}[ht]
    \centering
    \begin{tabular}{l l c l c c}
        \toprule
        \textsc{Model} & \textsc{Arch.} & \textsc{Tokens (B)} & \textsc{Teacher} & \textsc{MMLU Score} & \textsc{Relative Score (\%)} \\
        \midrule
        \textsc{SUPRA} & R & 100 & Mistral 7B & 34.2 & 24.6 \\
        \textsc{Mamba2-Llama 3} & R & 20 & Llama 3 8B & 43.2 & 43.8 \\
        \textsc{Mamba2-Llama 3} & H & 20 & Llama 3 8B & 56.7 & 76.2 \\
        \textsc{LoLCATs} & H & 0.04 & Mistral 7B & 51.4 & 70.5 \\
        \textsc{LoLCATs} & H & 0.04 & Llama 3 8B & 52.8 & 66.8 \\
        \textbf{\textsc{Llamba-8B}} & R & 12 & Llama 3.1 8B & \textbf{60.0} & \textbf{80.6} \\
        \bottomrule
    \end{tabular}
    \caption{MMLU performance of different models along with their respective architectures, training tokens, and teachers.
    Along with accuracy, each model is annotated with the number of training or distillation tokens (in trillions) and its architecture—Recurrent (R)
    or Hybrid (H).
    The relative score indicates the MMLU score relative to the teacher's score where the lower threshold is taken    
    to be 25 (random guessing). 
    }
\label{tab:comparison_against_distillation_methods}
\end{table}
    
\subsection{Throughput}
\label{subsec:throughput}
Llamba achieves higher throughput than its Llama-3.1-8B teacher (see \Cref{fig:throughput_over_batch_size}), even when Llama-3.1-8B generates four times fewer tokens.
This performance gain stems from Llamba’s recurrent Mamba-2 layers, whose state size remains constant regardless of sequence length.
Additionally, Llamba incorporates MLPs with fewer temporal mixing layers than \citet{mamba2}, enabling it to:
(1) scale to batches twice as large as a pure Mamba-2 model, as MLPs are stateless in time, and
(2) reduce kernel launch overhead when CUDA graph optimization is not applied, since Mamba layers typically require more kernel preparation time.

We have evaluated the throughput of Llama-3.1-8B and Llamba-8B models on a single NVIDIA H100 80GB HBM3. To ensure a fair comparison, all models were tested under a reasonable level of optimization, using \texttt{torch.compile(model, fullgraph=True)} and CUDA graph for consistent performance baselines.

\begin{figure}
    \centering
    \includegraphics[width=0.95\linewidth]
    {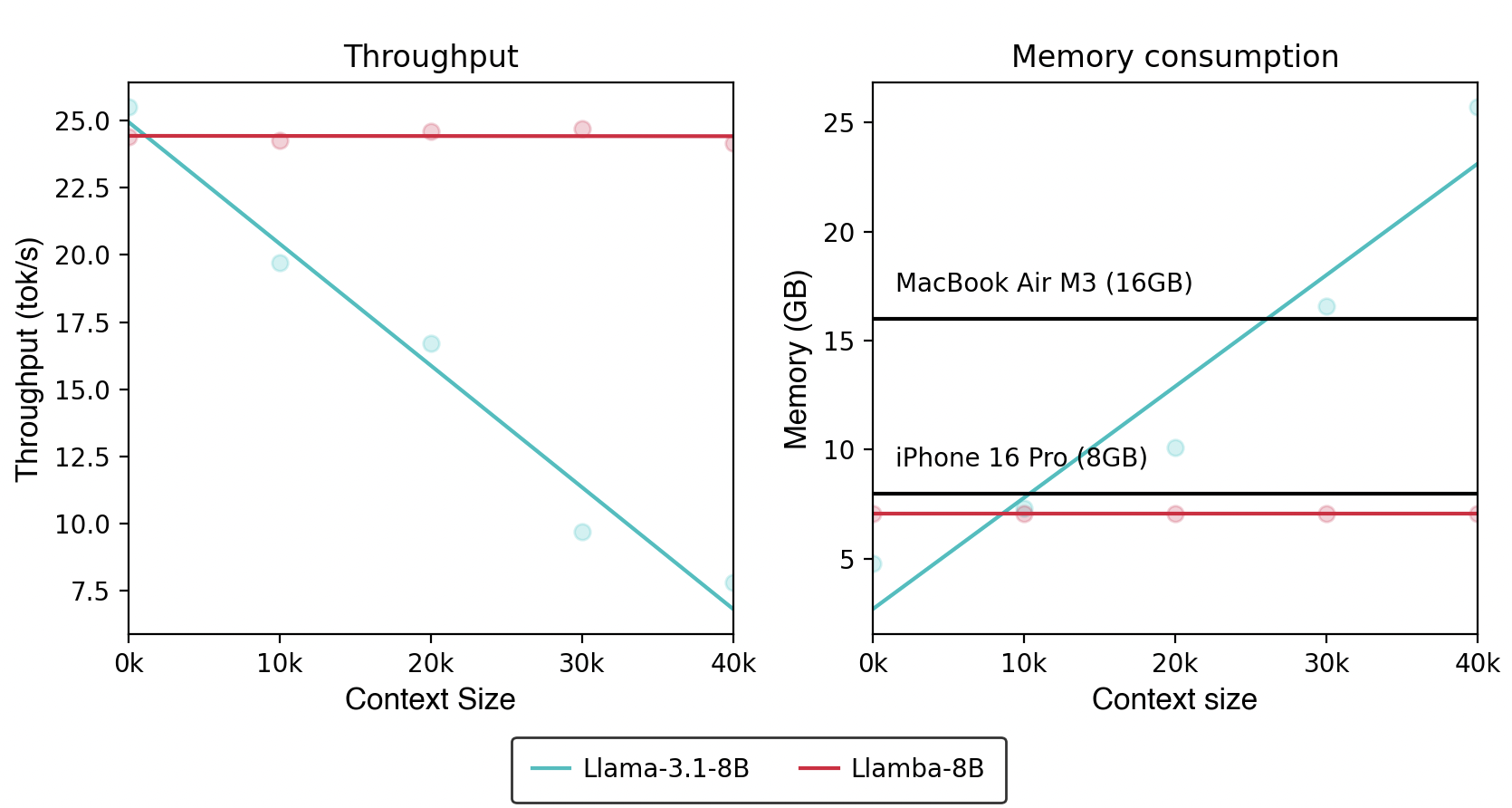}
    \caption{
    Comparison of on-device decoding throughput and memory consumption between Llamba-8B and Llama-3.1-8B at 4 bit quantization in MLX running on Apple Silicon M3 Pro (36GB). Llamba maintains constant high throughput and low memory consumption while the inference performance of Llama drops linearly with increasing context size.
    }
    \label{fig:on_device_comparison}
\end{figure}

Furthermore, on-device evaluation results highlight Llamba’s exceptional performance in decoding scenarios with constrained hardware. Specifically, on Apple Silicon M3 Pro (36GB) using MLX, Llamba maintains consistent high throughput and low memory consumption at 4-bit quantization (see \Cref{fig:on_device_comparison}). In contrast, the inference performance of Llama-3.1-8B deteriorates linearly with increasing context size, emphasizing the superior efficiency of Llamba in handling long sequences.

\section{Conclusion}

The Llamba model family represents a significant step forward in creating efficient and scalable recurrent language models. It achieves high performance with less than 0.1\% of the data typically required for similar models while maintaining strong performance across various benchmarks.

We see great promise in distilling pre-trained transformers into subquadratic architectures. Future directions include improving the quality and diversity of datasets used in distillation, optimizing the handling of long contexts, and expanding Llamba’s deployment to real-time, low-power applications such as IoT devices and wearable technology. Refining the distillation process further could unlock new capabilities and broaden the applications of this model family, solidifying its impact on efficient language modeling.

\bibliography{main}
\bibliographystyle{plainnat}

\appendix
\section{Comparison of Downstream Performance}
We evaluate a range of language models across multiple benchmarks, measuring their performance in both zero-shot and few-shot settings. ARC-Challenge and ARC-Easy \citep{arc}, and PIQA \citep{piqa} were evaluated with 0 and 25 shots, Winogrande \citep{winogrande} with 0 and 5 shots, HellaSwag \citep{hellaswag} and OpenBookQA \cite{OpenBookQA} with 0 and 10 shots, and Lambada \citep{lambada} and MMLU \citep{mmlu} with 0 and 5 shots. 
The results, reported in \Cref{tab:performance_comparison_1,tab:performance_comparison_2}, use normalized logits for ARC-Challenge, ARC-Easy, PIQA, HellaSwag, and OpenBookQA. These benchmarks cover a range of reasoning, commonsense, and language comprehension tasks, providing insight into the models’ ability to process different types of contextual dependencies.

Along with Llamba-1B, Llamba-3B, and Llamba-8B, we evaluate different quadratic and sub-quadratic architectures. Specifically, we include Transformer-based models such as Llama-3.2-1B-Instruct, Llama-3.2-3B-Instruct, and Llama-3.1-8B-Instruct \citep{llama}, as well as Qwen2.5-3B \citep{qwen2}. Models integrating recurrent components include Falcon-Mamba-7B-Instruct and Falcon3-Mamba-7B-Instruct \citep{falcon}, RecurrentGemma-2B-it and RecurrentGemma-9B-it \citep{recurrentgemma}, and Mamba varoatopms \citep{mamba1,mamba2}. 
Additionally, we evaluate Zamba2-7B-Instruct \citep{zamba2}, a hybrid incorporating both attention and SSMs. 
Whenever an instruct-tuned version is available, we use it, as these models generally perform better. These models differ in training strategies, data sources, and architectural choices, enabling a broad comparison of language model capabilities across different methodologies.

\begin{table}[H]
    \small
    \centering
    \setlength{\tabcolsep}{3.2pt}
    \caption{
    Comparison of downstream performance (accuracy \%) across various models in zero-shot and few-shot settings.
    For ARC-Challenge, ARC-Easy, and PIQA, we have used normalized logits' results.
    Along with accuracy, each model is annotated with the number of training or distillation tokens (in trillions) and its architecture—Recurrent (R), Transformer (T), or Hybrid (H). For models with a sliding window, the window size is also specified.
    We use an instruct-tuned version whenever one is available; however, we exclude this label for brevity.
    }
    \label{tab:performance_comparison_1}
    \begin{tabular}{l c c c c c c c c c c}
        \toprule
        \multirow{2}{*}{\textsc{Model}} 
         & \multirow{2}{*}{\textsc{Arch.}}
         & \multirow{2}{*}{\textsc{Tokens (T)}}
         & \multicolumn{2}{c}{\textsc{ARC Challenge}} 
         & \multicolumn{2}{c}{\textsc{ARC Easy}} 
         & \multicolumn{2}{c}{\textsc{PIQA}} 
         & \multicolumn{2}{c}{\textsc{Winogrande}} \\
        \cmidrule(r){4-5} \cmidrule(r){6-7} \cmidrule(r){8-9} \cmidrule(r){10-11}
         & & & 0-shot & 25-shot & 0-shot & 25-shot & 0-shot & 10-shot & 0-shot & 5-shot \\
         
        \midrule

        \textsc{Llama-3.2-1B (Teacher)} & T & $\leq 9$   & 38.1 & 42.0 & 68.5 & 71.8 & 74.4 & 75.4 & 59.7 & 62.0 \\
        \textbf{\textsc{Llamba-1B}}      & R & 0.008      & 37.2 & 41.8 & 69.5 & 71.2 & 74.0 & 74.3 & 60.6 & 58.1 \\
        \textsc{Mamba-1.4B}             & R & 0.3        & 32.9 & 36.0 & 60.9 & 66.6 & 73.7 & 74.4 & 60.6 & 60.1 \\
        \textsc{RecurrentGemma-2B}      & H ($w=2048$) & $\leq 2$  & 35.6 & 48.0 & 51.2 & 73.3 & 67.2 & 75.8 & 55.7 & 64.1 \\
        
        \midrule

        \textsc{Qwen2.5-3B}             & T & $\leq 18$  & 48.1 & 60.8 & 72.9 & 85.1 & 78.3 & 79.8 & 69.8 & 71.3 \\
        \textsc{Llama-3.2-3B (Teacher)} & T & 9          & 45.6 & 52.1 & 74.3 & 79.8 & 75.8 & 77.7 & 67.6 & 68.8 \\
        \textbf{\textsc{Llamba-3B}}      & R & 0.01       & 48.5 & 53.0 & 79.0 & 81.1 & 78.6 & 79.5 & 70.4 & 72.4 \\
        \textsc{Mamba2-2.8B}            & R & 0.3        & 35.9 & 39.5 & 64.3 & 71.8 & 75.6 & 76.4 & 63.4 & 64.6 \\
        
        \midrule

        \textsc{Qwen2.5-7B} & T & 18 & 55.1 & 67.0 & 81.3 & 89.5 & 80.3 & 82.4 & 71.1 & 75.1 \\
        \textsc{Llama-3.1-8B (Teacher)} & T & 15         & 55.1 & 60.0 & 81.7 & 85.8 & 81.1 & 82.4 & 73.9 & 77.3 \\
        \textbf{\textsc{Llamba-8B}}      & R & 0.012      & 54.6 & 60.0 & 82.5 & 85.8 & 80.9 & 81.5 & 73.3 & 76.9 \\
        \textsc{Falcon3-Mamba-7B}       & R & 7.3        & 53.2 & 65.9 & 72.5 & 86.7 & 79.7 & 82.3 & 69.1 & 72.1 \\
        \textsc{Zamba2-7B}             & H & 2.1        & 56.1 & 68.3 & 80.6 & 88.7 & 81.1 & 81.3 & 76.9 & 80.1 \\
        \textsc{RecurrentGemma-9B}      & H ($w=2048$) & 2       & 57.1 & 60.2 & 78.9 & 84.5 & 80.6 & 81.7 & 73.7 & 75.6 \\
        \bottomrule
        \multicolumn{11}{l}{} \\
    \end{tabular}
\end{table}

\begin{table}[H]
    \small
    \setlength{\tabcolsep}{3.5pt}
    \centering
    \caption{
    Comparison of downstream performance (accuracy \%) across various models in zero-shot and few-shot settings. For HellaSwag and OpenBookQA, we have used normalized logits' results.
    Along with accuracy, each model is annotated with the number of training or distillation tokens (in trillions) and its architecture—Recurrent (R), Transformer (T), or Hybrid (H). For models with a sliding window, the window size is also specified.
    We use an instruct-tuned version whenever one is available; however, we exclude this label for brevity.
    }
    \label{tab:performance_comparison_2}
    \begin{tabular}{l c c c c c c c c c c}
        \toprule
        \multirow{2}{*}{\textsc{Model}} 
         & \multirow{2}{*}{\textsc{Arch.}} 
         & \multirow{2}{*}{\textsc{Tokens (T)}}  
         & \multicolumn{2}{c}{\textsc{HellaSwag}} 
         & \multicolumn{2}{c}{\textsc{Lambada}} 
         & \multicolumn{2}{c}{\textsc{MMLU}} 
         & \multicolumn{2}{c}{\textsc{OpenBookQA}} \\
        \cmidrule(r){4-5} \cmidrule(r){6-7} \cmidrule(r){8-9} \cmidrule(r){10-11}
         & & & 0-shot & 10-shot & 0-shot & 10-shot & 0-shot & 5-shot & 0-shot & 10-shot \\
        \midrule
        \textsc{Llama-3.2-1B (Teacher)} & T & $\leq 9$    & 60.8 & 59.4 & 60.1 & 53.1 & 46.0 & 45.5 & 34.6 & 37.6 \\
        \textbf{\textsc{Llamba-1B}}      & R & 0.008       & 61.2 & 60.2 & 48.4 & 39.0 & 38.0 & 31.3 & 37.0 & 38.0 \\
        \textsc{Mamba-1.4B}             & R & 0.3         & 59.1 & 59.6 & 64.4 & 57.0 & 24.7 & 24.8 & 36.8 & 37.4 \\
        \textsc{RecurrentGemma-2B}      & H ($w=2048$) & $\leq 2$  & 60.3 & 69.4 & 52.5 & 53.0 & 40.2 & 42.1 & 30.4 & 43.2 \\
        
        \midrule

        \textsc{Qwen2.5-3B}             & T & $\leq 18$   & 74.9 & 75.2 & 65.8 & 58.1 & 65.5 & 66.4 & 41.8 & 46.2 \\
        \textsc{Llama-3.2-3B (Teacher)} & T & 9           & 70.4 & 73.2 & 65.9 & 61.9 & 60.4 & 59.8 & 35.8 & 39.6 \\
        \textbf{\textsc{Llamba-3B}}      & R & 0.01        & 73.8 & 74.3 & 65.8 & 60.0 & 52.7 & 50.3 & 42.8 & 42.8 \\
        \textsc{Mamba2-2.8B}            & R & 0.3         & 66.2 & 66.6 & 68.1 & 61.2 & 25.7 & 25.1 & 40.4 & 42.0 \\
        
        \midrule

        \textsc{Qwen2.5-7B} & T & 18 & 80.5 & 81.3 & 69.5 & 62.7 & 71.7 & 74.4 & 48.6 & 52.0 \\
        \textsc{Llama-3.1-8B (Teacher)} & T & 15          & 79.3 & 80.0 & 73.0 & 65.6 & 68.0 & 68.4 & 43.0 & 48.2 \\
        \textbf{\textsc{Llamba-8B}}      & R & 0.012       & 77.6 & 78.7 & 69.4 & 65.0 & 61.0 & 60.0 & 43.4 & 45.8 \\
        \textsc{Falcon3-Mamba-7B}       & R & 7.3         & 79.8 & 81.6 & 67.5 & 63.6 & 65.0 & 66.0 & 48.0 & 50.2 \\
        \textsc{Zamba2-7B}             & H & 2.1         & 81.5 & 83.5 & 74.6 & 68.6 & 64.7 & 67.2 & 45.2 & 52.4 \\
        \textsc{RecurrentGemma-9B}      & H ($w=2048$) & 2       & 80.1 & 80.9 & 54.1 & 69.6 & 55.1 & 56.5 & 46.0 & 49.2 \\
        \bottomrule
        \multicolumn{11}{l}{} \\
    \end{tabular}
\end{table}

\end{document}